\documentclass[11pt]{article}

\usepackage{acl}

\usepackage{times}
\usepackage{latexsym}

\usepackage[T1]{fontenc}

\usepackage[utf8]{inputenc}

\usepackage{microtype}

\usepackage{inconsolata}
\usepackage{microtype}
\usepackage{todonotes}
\usepackage{booktabs}
\usepackage{paralist}
\usepackage{colortbl}

\usepackage{xcolor}
\usepackage{hyperref}
 \definecolor{darkblue}{rgb}{0, 0, 0.5}
  \hypersetup{colorlinks=true, citecolor=darkblue, linkcolor=darkblue, urlcolor=darkblue}

\usepackage{xstring}

\usepackage{color}
\usepackage{graphicx}
\usepackage{tabularx}
\usepackage{soul}

\title{Are Sounds Sound for Phylogenetic Reconstruction?}

\author{
  Luise Häuser \\
  Heidelberg Institute for Theoretical Studies\\
  \texttt{luise.haeuser@h-its.org} \\
  \And
  Gerhard Jäger \\
  University of Tübingen \\
  \texttt{gerhard.jaeger@uni-tuebingen.de} \\
  \AND
  Taraka Rama \\
  Independent Researcher \\
  \texttt{taraka.kasi@gmail.com}
  \And
  Johann-Mattis List\\
  MPI-EVA / Univ. of Passau\\
  \texttt{mattis.list@uni-passau.de}\\
  \And
  Alexandros Stamatakis\\
  Institute of Computer Science\\ FORTH\\
  \texttt{stamatak@ics.forth.gr}
}

\begin{document}
\maketitle
\begin{abstract}
In traditional studies on language evolution, scholars often emphasize the importance of sound laws and sound correspondences for phylogenetic inference of language family trees. However, to date, computational approaches have typically not taken this potential into account.  Most computational studies still rely on lexical cognates as major data source for phylogenetic reconstruction in linguistics, although there do exist a few studies in which authors praise the benefits of comparing words at the level of sound sequences. Building on (a)~ten diverse datasets from different language families, and (b)~state-of-the-art methods for automated cognate and sound correspondence detection, we test, for the first time, the performance of sound-based versus cognate-based approaches to phylogenetic reconstruction.  Our results show that phylogenies reconstructed from lexical cognates are topologically closer, by approximately one third with respect to the generalized quartet distance on average, to the gold standard phylogenies than phylogenies reconstructed from sound correspondences.
\end{abstract}

\begin{figure*}[tb!]
  \centering
  \includegraphics[width=1\textwidth]{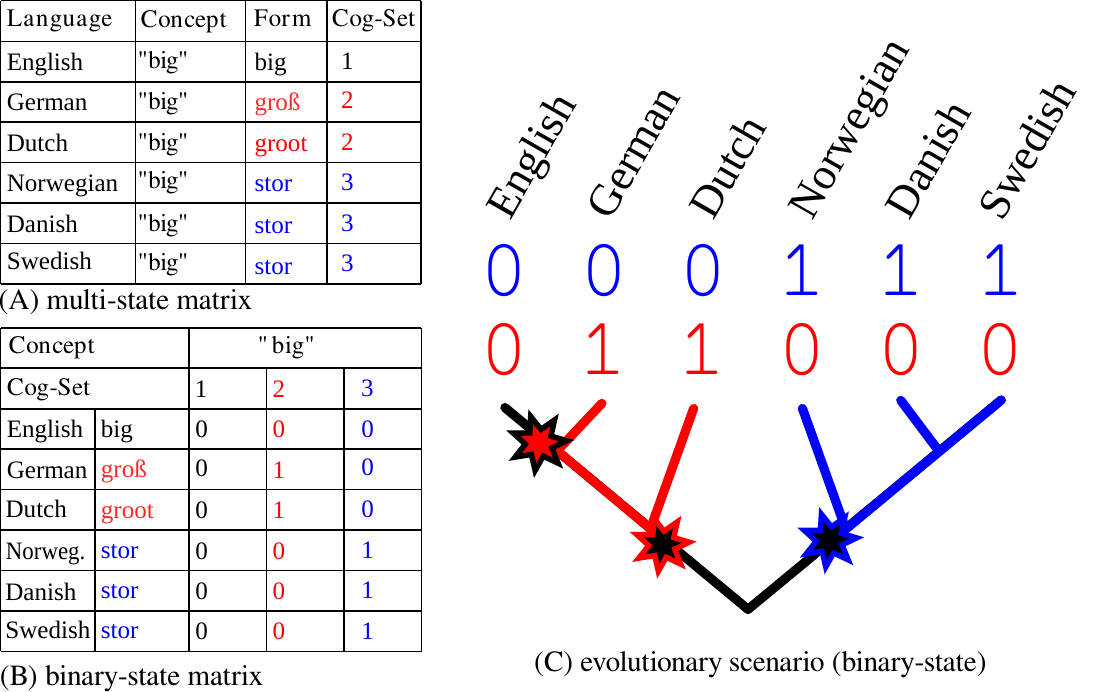}
  \caption{Gain-loss processes derived from binary cognate vectors. A shows a wordlist where
    cognate words are encoded as multi-state characters. B shows the corresponding binary encoding. C shows
    how gain and loss processes are modeled on a phylogenetic tree.}\label{fig:1}
\end{figure*}

\section{Introduction}

Although controversially discussed in the beginning \citep{Holm2007}, quantitative approaches to phylogenetic reconstruction based on Bayesian phylogenetic inference frameworks have now become broadly accepted and used in the field of comparative linguistics. This is reflected by the increasing number of computer-based phylogenies that have been proposed for the world's largest language families -- Dravidian \citep{Kolipakam2018}, Sino-Tibetan \citep{Sagart2019}, and Indo-European \citep{Heggarty2023} -- and even fully automated workflows, in which even cognate words are identified automatically, have shown to be comparatively robust \citep{rama2018automatic}.  While rarely practiced in the pre-computational past of historical linguistics, computing detailed, fully resolved phylogenies with branch lengths and at times even estimated divergence times, has now become a routine tasks in contemporary language evolution studies.

Although traditional scholars have started to accept computational language phylogenies as a new tool deserving its place in the large tool chain of comparative linguistics, scholars still express substantial skepticism against most language phylogenies that have been inferred so far. One of the major arguments typically mentioned in this context is that phylogenetic approaches are usually based on cognate sets (sets of historically related words) that are identified in semantically aligned word lists. Since these \emph{cognate sets} reflect \emph{lexical data} only, many scholars mistrust them, given that lexical data are assumed to be substantially less stable over time than other aspects of languages~\cite{CampbellPoser2008}. Yet, for being able to infer stable phylogenetic trees a mix of conserved characters and more variable characters might be more beneficial.

In classical historical linguistics, the data used for subgrouping are traditionally composed of small collections of so-called \emph{shared innovations} \citep{Dyen1953}. What counts as a shared innovation has itself never been clearly defined in the literature, but the largest amount of data used by scholars is traditionally taken from sound correspondences or supposed sound change processes (compare, for example the data in \citealt[305]{Anttila1972}). Although it is controversially debated in the field \citep{Ringe2002,Dybo2008}, many classical linguists still emphasize that sound correspondences are largely superior to lexical data to determine subgrouping.

There have only been few attempts to assess how well quantitative approaches to phylogenetic reconstruction perform when using sound correspondences instead of lexical cognates \citep{Chacon2015a}. The main reason is that encoding data to compute phylogenies from sound change patterns is tedious and labour-intensive even for a dataset comprising only 20 languages.  Therefore, there have been but a few attempts to assess how well quantitative approaches to phylogenetic reconstruction perform when using sound correspondences instead of lexical cognates.

Here we build on state-of-the-art methods for automatic cognate detection and phonetic alignment in historical linguistics \citep{List2016g} and combine them with novel approaches for inferring sound correspondence patterns in multilingual datasets \citep{List2019a}. Using this machinery we have devised a new workflow for phylogenetic reconstruction based on sound correspondence patterns. With a new collection of ten gold standard datasets, we test our workflow and compare it with alternative workflows that are exclusively based on lexical data. Our results indicate that sound correspondence patterns are substantially less suitable for the purpose of computer-based phylogenetic reconstruction than postulated.




\begin{table*}[!tb]
\centering
\resizebox{\textwidth}{!}{%
\begin{tabular}{lrrrrrrr}
\toprule
 \textbf{Dataset}      &   \textbf{Words} &   \textbf{Concepts} &   \textbf{Languages} &
 \textbf{Distances} &   \textbf{Sounds} &   \textbf{Word Length} &  \\
\midrule
 ConstenlaChibchan (Constenla Umaña 2005)        & 1214 & 106 & 24 & 0.1  & 21.71 & 3.86 \\
 CrossAndean (Blum et al. 2023)                & 2637 & 150 & 19 & 0.03 & 28.89 & 4.32 \\
 Dravlex (Kolipakam et al. 2018)                  & 1341 & 100 & 20 & 0.06 & 36.85 & 4.53 \\
 FelekeSemitic (Feleke 2021)            & 2412 & 150 & 19 & 0.05 & 45.32 & 4.99 \\
 Hattorijaponic (Hattori 1973)          & 1710 & 197 & 10 & 0.03 & 34.9  & 4.47 \\
 HouChinese (Ho\'u 2004)                     & 1816 & 139 & 15 & 0.05 & 43    & 6.21 \\
 LeeKoreanic (Lee 2015)                  & 1960 & 205 & 14 & 0.01 & 36.93 & 4.31 \\
 RobinsonAP (Robinson and Holton 2012)                & 1424 & 216 & 13 & 0.03 & 24.38 & 4.51 \\
 WalworthPolynesian (Walworth 2018)        & 6113 & 207 & 31 & 0.05 & 21.03 & 4.51 \\
 ZhivlovObugrian (Zhivlov 2011)        & 1879 & 110 & 20 & 0.04 & 32.65 & 3.65 \\
\bottomrule
\end{tabular}}
\caption{Datasets and general aspects of the data. Distances refer to the average pairwise distance between all language pairs in the sample, derived from shared cognate counts (using the LingPy software). Number of sounds refers to the number of distinct sounds per language (on average), and the word length refers to the average length of the words observed in each dataset.}
\label{tab:data}
\end{table*}
\nocite{constenla2005,crossandean,Kolipakam2018,felekesemitic,hattorijaponic,hou2004,leekorean,Robinson2012,Walworth2018,zhivlovobugrian}

\section{Background}\label{sec:motiv}

The majority of previous work on phylogenetic reconstruction using Bayesian phylogenetic inference~\cite{Kolipakam2018,Sagart2019,rama2018automatic} is based on cognate sets that are encoded as binary vectors. The presence or absence of a language in a cognate set is thus encoded as \textbf{1} or \textbf{0}, respectively.
Subsequently, phylogenetic trees are inferred by assuming that cognate sets evolve along a phylogenetic tree via a gain and loss processes (see Figure~\ref{fig:1}).

The binary-state encoding is the most frequently used encoding technique; we deploy it in this study as well. Once such a dataset has been assembled, binary state data evolution can be modeled via a time-reversible binary state Continuous Time Markov Chain model (\emph{binary-CTMC}, \citealt{bouckaert2012mapping}), which allows for gain and loss events to occur for an arbitrary number of times. Branch lengths on these trees reflect the mean number of expected substitutions (gain/loss events) per binary character site. 

The \textit{major contributions} of this study are:
(1)~We provide an automated workflow that allows to infer cognates and correspondence patterns and analyze them with the help of Bayesian phylogenetic inference methods, (2)~we cross-validate the Bayesian inference results via Maximum Likelihood (ML) tree reconstructions and thereby discover that default Bayesian priors that typically work well on molecular data can induce a prior bias when analyzing language datasets, (3) we show how the quality of phylogenetic reconstruction approaches based on sound correspondences can be compared to phylogenetic reconstruction based on lexical data, and in this way, and (4)~we put the debate about the usefulness of sound-based as opposed to cognate-based phylogenies to the test.

As an early example for sound-based approaches to phylogenetic reconstruction, \citet{hruschka2015detecting} apply a CTMC model that allows for transitions between a fixed number of sounds for detecting the important sound changes in a dataset comprising etymologies across Turkic languages. Hruschka et al.~do not infer phylogenies from their data. Instead, they use an established phylogeny (such established phylogenies are not readily available for many language families of the world) to infer branch lengths and transition probabilities between sounds in their data in order to detect sound changes at different time points in a time-calibrated family tree of Turkic.

\citet{Wheeler2015b} start from typical word lists (that would otherwise be used in phylogenetic reconstruction based on lexical data) and apply a parsimony-based algorithm that aligns words regardless if they are cognate or not, reconstructs a hypothetical ancestral word from the alignment, and seeks to infer the phylogeny that explains the observed sequences via the minimum amount of changes/mutations \citep{Sankoff1975}. In a later study, \citet{Whiteley2019} apply the same approach to a dataset of Bantu languages. The method by \citet{Wheeler2015b} is linguistically debatable, since words are not assigned to cognate sets prior to aligning them. It is well known that there is a strict difference between regular sound change processes and processes resulting from lexical replacement \citep{Hall2010} and that even words that are cognate are not necessarily fully \emph{alignable}
\citep[10]{Schweikhard2020}.

\citet{Chacon2015a} start from manually extracted sound correspondence patterns for consonants in a dataset of 21 Tukanoan languages, to which proto-forms had also been manually added. Based on these sound correspondence patterns, they apply---in analogy to \citet{Wheeler2015b}---an algorithm that searches for the tree that provides the most parsimonious explanation for sound evolution. In contrast to \citet{Wheeler2015b}, however, they added specific constraints for the transitions from one sound to another sound, which were based on expert judgments for the Tukanoan language family. The approach by \citet{Chacon2015a}, finally, requires an enormous amount of preprocessing that entails the risk of inducing circular results, since proto-forms and major directions of sound change processes are required to be known in advance. While all approaches exhibit individual shortcomings, one of the largest shortcomings lies in the fact that it is very difficult to apply them systematically. This is also supported by the observation that no additional analogous studies have been conducted by other teams, despite the fact that all of the above methods have been proposed years ago.

\section{Materials and Methods}

\subsection{Materials}
In order to test whether sound correspondence patterns improve phylogenetic reconstruction or not, we selected ten datasets from the Lexibank repository (\url{https://lexibank.clld.org}, \citealt{List2022e}) which were previously used to investigate the regularity of correspondence patterns in comparative cognate-coded wordlists \citep{Blum2023}. Lexibank offers published datasets in standardized formats (so-called Cross-Linguistic Data Formats, see \citealt{Forkel2018a}). According to these standards, languages are linked to the Glottolog reference catalog (offering access to expert phylogenies and geolocations, \url{https://glottolog.org}, \citealt{Glottolog}), concepts are linked to the Concepticon reference catalog (offering fundamental definitions of semantic glosses and further information on concept properties, \url{https://concepticon.clld.org}, \url{Concepticon}), and sounds are provided in the phonetic transcription underlying the Cross-Linguistic Transcription Systems initiative (a reference catalog on speech sounds, offering a dynamic system that defines transcriptions for more than 8000 standard speech sounds observed in linguistic datasets, \url{https://clts.clld.org}, \citealt{CLTS,Anderson2018}).

Data were preprocessed by first computing the phonetic alignment of all cognate sets in the data using the multiple alignment method proposed by \citet{List2014d}. In a second step, these alignments were automatically \emph{trimmed}, using the method proposed by \citet{Blum2023}, which identifies alignment columns with many gaps and ignores them, assuming that these result from morphological variation that would confuse cognate judgments.For phylogenetic inferences on molecular sequence data \citet{tan2015current} suggest that filtering worsens phylogenetic inference accuracy. The study by \citet{Blum2023}, however, shows that -- for linguistic data -- the overall regularity among cognates increases substantially, when trimming alignments systematically. Since regular sound correspondences provide the basis for the identification of classical sound laws that linguists typically use for the traditional subgrouping by shared innovations, we therefore consider the use of trimmed data as advantageous over using untrimmed alignments. Using trimmed phylogenies also has the advantage of reducing the noise, as can be seen from a rather drastic drop in the number of divergent sites in phylogenetic datasets that have been trimmed. However, it is beyond doubt that a closer investigation of the effects of trimming should be carried out in follow-up studies.
In a third step, the method by \citet{List2019a} was used to compute correspondence patterns of the data. Phonetic alignments were conducted with LingPy (2.6.11, \citealt{LingPy}, \url{https://pypi.org/project/lingpy}). Trimming and correspondence pattern detection were carried out with LingRex (1.4.1, \citealt{LingRex}, \url{https://pypi.org/project/lingrex}).

Having identified correspondence patterns from the data, both the information on cognate sets and the information on correspondence patterns were converted into binary presence-absence matrices in Nexus format \citep{Maddison1997}, suitable for subsequent phylogenetic analysis.

\subsection{Methods}

Different methods for phylogenetic reconstruction have been described in the literature and have been controversially discussed among scholars for some time. Here we test two very basic approaches, Bayesian Inference and Maximum Likelihood. Since the data that we use for the inference of phylogenies comes in two flavors, derived as binary presence-absence matrices from cognate sets and from sound correspondence patterns, we test the methods on three different \emph{character matrices}, namely the \emph{cognate matrix}, derived from cognate judgments, the \emph{sound correspondence matrix}, derived from sound correspondence patterns, and a \emph{combined matrix}, in which we combine (concatenate) the cognate and the character matrix within a single new matrix.
 
In our experiments, we test three basic hypotheses. The first hypothesis assumes that phylogenetic inference on cognate sets is more accurate than phylogenetic inference based on sound correspondence patterns. The second hypothesis assumes that phylogenetic inference based on sound correspondence patterns is more accurate than phylogenetic inference based on cognate sets. The third hypothesis assumes that both character types do not differ substantially regarding their phylogenetic signal.
\subsubsection{Bayesian Inference}

Phylogenetic inferences were conducted using \emph{MrBayes} \citep{mrbayes3}, version 3.2.7. For the final inferences presented here we used the following prior settings for all datasets:
(1)~\(\mathrm{Dirichlet}(1.0, 1.0)\) prior for base frequencies, (2)~gamma-distributed rates, approximated by 4 discrete categories, with standard exponential prior for the shape of the gamma distribution that models among site rate heterogeneity,
(3)~uniform prior over tree topologies, and
(4)~strict clock model of branch lengths.
 
We initially used an an exponential distribution with a rate of \(1.0\) as prior for the $\Gamma$ model of rate heterogeneity. This prior constrains the $\alpha$ shape parameter of the $\Gamma$ distribution to relatively small values. 
As a consequence, MrBayes obtains $\alpha$ values below $10$ for almost all data sets and posterior samples it draws. This indicates a high to moderate degree of rate heterogeneity.
However, our ML analyses (see below) yielded substantially higher ML estimates for $\alpha$ on some datasets. 
To investigate this discrepancy, we repeated the Bayesian inferences, now using \(\mathrm{Uniform}(0.01, 100)\) as a prior for the $\Gamma$ distribution of rate heterogeneity. As a consequence, we obtained a different distribution of the $\alpha$ values that better reflects the corresponding ML estimates. 

The more informative default exponential prior in MrBayes has presumably been developed for molecular datasets, which usually exhibit a high degree of rate heterogeneity. In other words, ML estimates of $\alpha$ exhibit a small variance (see, e.g., \url{https://github.com/angtft/RAxMLGroveScripts/blob/main/figures/test_ALPHA.png} and the corresponding paper by \citet{hoehler2021grove}). When executing inferences on language datasets, using this default molecular prior can hence bias the results. That is, had we not conducted complimentary ML analyses, this surprising dataset-dependent bi-modal distribution of $\alpha$ values on language datasets (see Table~\ref{tab:2}) would have gone unnoticed. We thus strongly advocate that all default priors for molecular datasets should be carefully and critically re-assessed when conducting Bayesian inferences on language datasets and that ML analyses should always complement Bayesian Inferences.
\begin{table*}[tb!]
  \centering
  \begin{tabular}{lccc}
    \toprule
    \textbf{Dataset}   & \textbf{Cognates} & \textbf{Sound Correspondences} & \textbf{Combined} \\\midrule
 ConstenlaChibchan  & 0.592           & \textbf{99.871} & 4.178 \\
 CrossAndean          & 1.243           & 6.334           & 1.154 \\
 Dravlex            & 0.702           & 4.301           & 2.234 \\
 FelekeSemitic      & 1.062           & 7.430           & 2.693 \\
 Hattorijaponic    & \textbf{99.848} & \textbf{99.897} & \textbf{99.890} \\
 HouChinese               & 2.357           & 6.120           & 4.195 \\
 LeeKoreanic            & 8.316           & 8.420           & 3.284 \\
 RobinsonAP          & \textbf{99.869} & 15.269          & 3.486 \\
 WalworthPolynesian  & 1.333           & 4.233           & 1.624 \\
 ZhivlovObugrian  & \textbf{99.850} & 4.244           & 3.134
 \\\bottomrule
  \end{tabular}\caption{ML estimates of the alpha shape value of the Gamma model for among site rate heterogeneity for all languages and all character matrices. Values indicating an extremely low rate heterogeneity (all sites evolve at the same rate) are highlighted in bold.}%
  \label{tab:2}
\end{table*}

Motivated by this observation, the Bayesian analysis was repeated, now using a uniform prior over the interval $[0.01, 100.0]$ for $\alpha$.

We sampled the state of the Markov chain every 1,000th generation. We stopped MCMC chains when the average standard deviation of split frequencies (ASDSF) was below \(0.01\) after discarding the first \(25\% \) of the samples.\footnote{Note that convergence diagnosis metrics such as ASDSF can only serve to diagnose the failure of an MCMC chain to converge, but can never confirm its convergence.}

The median posterior value for $\alpha$ are shown in Table \ref{tab:2b}.
\begin{table*}[tb!]
  \centering
  \begin{tabular}{lccc}
    \toprule
    \textbf{Dataset}   & \textbf{Cognates} & \textbf{Sound Correspondences} & \textbf{Combined} \\\midrule
    ConstenlaChibchan & 1.758 & \textbf{53.115} & 1.138 \\
    CrossAndean & 1.620 & 19.558 & 0.400 \\
    Dravlex & 0.749 & 23.613 & 0.814 \\
    FelekeSemitic & 0.932 & 41.669 & 0.727 \\
    HattoriJaponic & \textbf{58.012} & \textbf{60.602} & 0.268 \\
    HouChinese & 3.011 & 27.476 & 0.933 \\
    LeeKoreanic & \textbf{52.045} & 39.354 & 0.058 \\
    RobinsonAP & \textbf{56.928} & \textbf{51.818} & 0.373 \\
    WalworthPolynesian & 1.480 & 4.348 & 0.800 \\
    ZhivlovObugrian & \textbf{58.652} & \textbf{51.280} & 0.507 \\
    \bottomrule
  \end{tabular}\caption{Median Bayesian estimates of the alpha shape value of the Gamma model for among site rate heterogeneity for all languages and all character matrices. Values indicating an extremely low rate heterogeneity (all sites evolve at the same rate) are highlighted in bold.}%
  \label{tab:2b}
\end{table*}
From the remaining \(75\% \) of the recorded samples from the two cold chains, 1,000 trees were drawn at random and used for further evaluation.




If one of the two individual character types provides the best results, this would be evidence for Hypothesis 1 or Hypothesis 2. If the combined dataset provides the best results, this would be evidence for Hypothesis 3.

To evaluate the quality of the inferred phylogenies, we used the classifications from Glottolog \citep{Glottolog}. The topological distance or degree of consistency of an inferred strictly binary (fully bifurcating) phylogeny and a (potentially polytomous/multi-furcating) Glottolog tree was measured as the \emph{generalized quartet distance} (GQD), as proposed in \citep{pompei2011accuracy}.\footnote{The GQD is a generalization of the well-known \emph{quartet distance} \citep{estabrook1985comparison} that allows to compare fully bifurcating trees with multi-furcating trees. The GQD is defined as the number of quartets that are not shared between the two trees, divided by the number of all possible quartets. The GQD is a number between 0 and 1, where 0 means that the two trees are identical, and 1 means that the two trees are completely different.}

\subsubsection{Maximum Likelihood Tree Inferences}
To exclude any potential bias by the selected tree inference method, we also conducted independent Maximum Likelihood (ML) tree inferences. 
For ML tree inference we used RAxML-NG \cite{kozlov2019raxmlng}, version 1.2.0. 
For each dataset and character matrix type (cognate/sound/concatenated) we 
executed 20 independent ML tree searches using the default tree search configuration of RAxML-NG (10 searches starting from random trees and 10 searches starting from randomized stepwise addition order parsimony trees) under the BIN+G model of binary character substitution with ML estimated base frequencies.
We approximate the $\Gamma$ model of rate heterogeneity via four discrete rates. Thus, each inference includes the ML estimate of the $\alpha \in [0.0201, 100]$ shape parameter that determines the shape of the $\Gamma$ distribution. 
The smaller the estimate of $\alpha$, the higher the rate heterogeneity in the respective dataset will be \cite{yang1995gamma}. 
For three matrices containing cognate data and for two matrices encoding sound correspondences, we obtain an estimate for $\alpha > 99.8$, which means that all sites evolve at the same rate and that there is essentially no rate heterogeneity. 
Hence, trees on these datasets could also be inferred without correcting for rate heterogeneity. 
For the remaining datasets, the ML estimates of $\alpha$ are below $20$, indicating a moderate to high degree of rate heterogeneity.
This extreme bi-modal distribution of $\alpha$ estimates differs substantially from the distribution we observe on tens of thousands of empirical (i.e., non-simulated) molecular datasets \cite{hoehler2021grove} (also see the respective distribution of $alpha$ values plot at \url{https://github.com/angtft/RAxMLGroveScripts/blob/main/figures/test_ALPHA.png} where $alpha$ values range approximately between 0.01 and 1.5).

We have currently not been able to identify which intrinsic dataset properties cause this surprising and extreme bi-modal distribution of $alpha$ values in language datasets. For the given datasets, we examined the number of concepts and languages under study, the dimensions of the MSAs and the average branch lengths in the trees inferred.
We determined the difficulty score using Pythia \cite{haag2022pythia} and the number of species using the method for species delimitation implemented in mPTP \cite{kapli2017mptp}. For none of these properties we were able to find a clear connection to the value estimated for $\alpha$. One path to explore in analogy to molecular biology is whether some language datasets should be regarded as representing individuals from a population of the same species while others represent distinct species. In fact, for molecular data we have thus far only observed such high estimates of $\alpha$ values (i.e., low or no rate heterogeneity) for population genetic datasets comprising sequences of individuals of the same species or closely related sub-species. 

\begin{table*}[tb]
  \centering
  \begin{tabular}{lccc}
    \toprule
    \textbf{Dataset}   & \textbf{Cognates} & \textbf{Sound Correspondences} & \textbf{Concatenated} \\\midrule
    ConstenlaChibchan  & 0.245                    & 0.414                          & \textbf{0.212}    \\
    CrossAndean        & \textbf{0.148}           & 0.523                          & 0.189             \\
    Dravlex            & 0.336                    & 0.351                          & \textbf{0.320}    \\
    FelekeSemitic      & \textbf{0.083}           & 0.146                          & 0.113             \\
    HattoriJaponic     & 0.585                    & 0.431                          & \textbf{0.362}    \\
    HouChinese         & \textbf{0.240}           & 0.494                          & 0.377             \\
    LeeKoreanic        & 0.224                    & 0.358                          & \textbf{0.157}    \\
    RobinsonAP         & 0.424                    & 0.281                          & \textbf{0.259}    \\
    WalworthPolynesian & 0.179                    & 0.252                          & \textbf{0.146}    \\
    ZhivlovObugrian    & 0.330                    & 0.356                          & \textbf{0.316}    \\\midrule
    \emph{median}      & 0.251                    & 0.358                          & \textbf{0.240}    \\\bottomrule
  \end{tabular}\caption{Generalized quartet distances (posterior medians) for Bayesian inference. The best result for each dataset is highlighted in bold.}%
  \label{tab:3}
\end{table*}

\begin{table*}[tb]
  \centering
  \begin{tabular}{lccc}
    \toprule
    \textbf{Dataset}   & \textbf{Cognates} & \textbf{Sound Correspondences} & \textbf{Combined} \\\midrule
    ConstenlaChibchan  &             0.335 &             0.360 &     \textbf{0.283} \\
    CrossAndean       &             0.246 &             0.470 &    \textbf{0.088} \\
    Dravlex       &             0.358 &             0.472 &    \textbf{0.307} \\
    FelekeSemitic      &             0.126 &    \textbf{0.103} &             0.126 \\
    HattoriJaponic     &    \textbf{0.532} &             0.681 &             0.559 \\
    HouChinese         &             0.224 &             0.529 &    \textbf{0.186} \\
    LeeKoreanic     &    \textbf{0.178} &             0.386 &             0.204 \\
    RobinsonAP         &             0.355 &    \textbf{0.321} &             0.348 \\
    WalworthPolynesian  &    \textbf{0.139} &             0.188 &             0.192 \\
    ZhivlovObugrian     &    \textbf{0.322} &             0.356 &             0.360  
    \\\midrule
    \emph{median}      &             0.284 &             0.373 &     \textbf{0.243}    \\\bottomrule
  \end{tabular}\caption{Generalized quartet distances between the gold standard trees and the the best-scoring ML tree inferred under the \textbf{BIN+G} model. The best result for each dataset is highlighted in bold.}%
  \label{tab:4}
\end{table*}

\subsection{Implementation}
Methods for data handling and preprocessing are implemented in Python (with specific requirements and software packages indicated above), R and Julia. For the phylogenetic analyses, dedicated third party packages are used. All information on how to replicate our study and how to inspect individual analyses are provided in the supplementary material accompanying this study. 

\section{Results}
\subsection{Bayesian Inference}

We computed the GQD to the goldstandard tree for each of the 1,000 samples from the posterior and computed the median for each dataset and character type. The results of our evaluation are shown in Table~\ref{tab:3}. As can be seen from the table, phylogenetic inferences based on cognate class data and on concatenated cognate/sound data provide results that are about equally good, with a slight advantage for concatenated data. Phylogenetic inference based on sound correspondences alone yields results that are clearly worse. The concatenated dataset provides the best results for seven out of ten datasets, while in three cases, the cognate class dataset provides the best results. The sound correspondence dataset never yields the best results.
These results provide clear evidence in favor of Hypothesis 1 and against Hypothesis 2. The decision about Hypothesis 3 is somewhat equivocal.

\subsection{Maximum Likelihood}

Table \ref{tab:4} shows the evaluation results for the ML based inferences. Note that we obtain slightly different results when calculate the average distance to all 20 inferred ML trees or when using the BIN model (without accounting for among site rate heterogeneity). The corresponding GQ distances can differ by up to 0.17, although differences of $> 0.05$ only occur for 7 of the 30 MSAs under study.
However, the following observations apply in all cases.
First, there is no dataset where the tree inferred on the sound correspondences is substantially closer to the gold standard than the trees inferred on the cognate or concatenated data.
On the other hand, there are three datasets (CrossAndean, HouChinese, LeeKoreanic) for which inferences on sound correspondence data yield trees with a substantially higher GQ distance to the gold standard.
Inferences on the cognate and combined datasets yield comparable distances to the gold standard. 
Hence, the results of our ML analyses are consistent with the Bayesian inference results.

\section{Discussion and Conclusion}

While our results are less conclusive than one
might expect, we think that they show clearly
enough that sound-correspondence-based phylogenies should be
taken with care. As we show, sound-correspondence-based phylogenies do rarely substantially outperform cognate-based phylogenies. Instead, we observe that cognate-based phylogenies are topologically much closer
to the gold standard on average. At this point, we cannot say, whether combined approaches significantly outperform phylogenies purely inferred from cognate sets. Future studies that expand the data we used in this study are needed to clarify this question.
 
Given the prior bias we observed for default parameter priors that work well for Bayesian inference on molecular data, we advocate for a critical re-assessment of all priors that are being routinely used in Bayesian analyses of language data. This re-assessment can be conducted by routinely executing analogous ML analyses and carefully inspecting all ML parameter estimates (branch lengths, tree length, base frequencies, $\alpha$ shape parameter) and not only focusing on the resulting tree topology. We also cross-checked the estimates for the base frequencies, but did not observe any discrepancies between ML and Bayesian Inference as a flat default ($\beta(1,1)$) prior was used. 
The reasons for the extreme bi-modal distribution of $\alpha$ values we observed remain unclear, despite the fact that we have assessed 30 different dataset characteristics and summary statistics that are, however, all uncorrelated with the $\alpha$ estimate. Using machine learning techniques to predict $\alpha$ values for datasets and thereby potentially understand the dataset properties responsible for this bi-modal distribution is not feasible due to an insufficient amount of available data. Investigating this issue hence remains subject of future work.

\section*{Supplementary Material}
The supplementary material including data and code necessary to replicate the experiments discussed in this study along with instructions on how to run the code are curated on GitHub (\url{https://github.com/lingpy/are-sounds-sound-paper}) and archived with Zenodo (\url{https://doi.org/10.5281/zenodo.10610428}). 

\section*{Limitations}
The ongoing debate of what evidence phylogenetic reconstruction should be based on cannot be considered as conclusively solved with this study, although we are confident that our contribution merits the attention of all scholars participating in the debate. One crucial weakness of our approach, which we cannot overcome completely at the moment, is the way we operationalize ``sound laws as evidence for phylogenetic reconstruction''. Here, we use sound correspondence patterns which we infer automatically from the data sets. One may criticize that this procedure is not identical with the way in which experts do cladistic subgrouping. In response to such criticism, we emphasize, however, that every attempt to arrive at a useful way to compare evidence based on sound correspondence patterns (and sound laws) with evidence based on cognate sets, must start at some point, and that we are convinced that this approach comes quite close to the evidence traditional scholars defending phylogenetic reconstruction by innovation have in mind.

\section*{Acknowledgments}

This research was supported by the Max Planck Society Research Grant \emph{CALC³} (JML, \url{https://digling.org}), 
the ERC Consolidator Grant \emph{ProduSemy} (JML, Grant No. 101044282, see \url{https://doi.org/10.3030/101044282}), the ERC Advanced Grant \emph{CrossLingFerence} (GJ, Grant. No. 834050, see \url{https://doi.org/10.3030/834050}), the Klaus-Tschira Foundation, and by the European Union (EU) under Grant
Agreement No 101087081 (AS, Comp-Biodiv-GR, see \url{https://doi.org/10.3030/101087081}). Views and opinions expressed are however those of the authors only and do not necessarily reflect those of the European
Union or the European Research Council Executive Agency (nor any other funding agencies involved). Neither the European Union nor the granting authority can be held responsible for them. We thank Maria Heitmeier and Harald Baayen for their valuable input regarding the computational models used in this study. 

\begin{center}
    \includegraphics[width=0.3\textwidth]{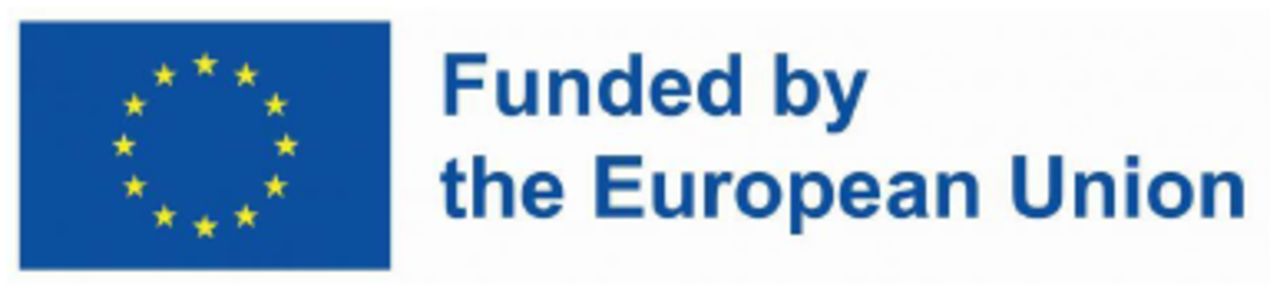}     
    \end{center}

\newpage

\end{document}